\documentclass[letterpaper]{article} 
\usepackage{aaai24}  
\usepackage{times}  
\usepackage{helvet}  
\usepackage{courier}  
\usepackage[hyphens]{url}  
\usepackage{graphicx} 
\usepackage{natbib}  
\usepackage{caption} 
\usepackage{algorithm}
\usepackage{algorithmic}

\usepackage{newfloat}
\usepackage{listings}
\DeclareCaptionStyle{ruled}{labelfont=normalfont,labelsep=colon,strut=off} 
\floatstyle{ruled}
\newfloat{listing}{tb}{lst}{}
\floatname{listing}{Listing}
\nocopyright
\usepackage{amsmath}
\usepackage{amssymb}
\usepackage{mathrsfs}
\usepackage{tabularx}
\usepackage{multirow}
\usepackage{makecell}
\usepackage{enumitem}
\usepackage{booktabs}

\newtheorem{definition}{Definition}

\title{CUTS+: High-dimensional Causal Discovery from Irregular Time-series}
\author{
Yuxiao Cheng\textsuperscript{1}\footnotemark[1]~~~ Lianglong Li\textsuperscript{1}\thanks{Equal Contribution}~~~ Tingxiong Xiao\textsuperscript{1}~~~ Zongren Li\textsuperscript{3}~~~ Qin Zhong\textsuperscript{3} \\
\textbf{Jinli Suo\textsuperscript{12}\footnotemark[2]~~~ Kunlun He
\textsuperscript{3}\footnotemark[2]~~~ 
}
}

\affiliations{
\textsuperscript{1}Department of Automation, Tsinghua University\\
\textsuperscript{2}Institute for Brain and Cognitive Science, Tsinghua University (THUIBCS)\\
\textsuperscript{3}Chinese PLA General Hospital\\
}

\begin{document}

\maketitle

\begin{abstract}
Causal discovery in time-series is a fundamental problem in the machine learning community, enabling causal reasoning and decision-making in complex scenarios. Recently, researchers successfully discover causality by combining neural networks with Granger causality, but their performances degrade largely when encountering high-dimensional data because of the highly redundant network design and huge causal graphs. Moreover, the missing entries in the observations further hamper the causal structural learning. To overcome these limitations, we propose CUTS+, which is built on the Granger-causality-based causal discovery method CUTS and raises the scalability by introducing a technique called coarse-to-fine-discovery (C2FD) and leveraging a message-passing-based graph neural network (MPGNN).  Compared to previous methods on simulated, quasi-real, and real datasets, we show that CUTS+ largely improves the causal discovery performance on high-dimensional data with different types of irregular sampling. 
\end{abstract}

\section{Introduction}

Analyzing complex interactions behind the observed time-series, i.e., time-series analysis, is a fundamental problem in machine learning and holds great potential in various real-world applications. However, revealing the complex relationships buried under massive amounts of variables can be challenging for algorithm design. Recently, approaches have been proposed to extract causal relationships from observational data \cite{tankNeuralGrangerCausality2022, loweAmortizedCausalDiscovery2022, khannaEconomyStatisticalRecurrent2020, rungeDetectingQuantifyingCausal2019, chengCUTSNeuralCausal2023, xuScalableCausalGraph2019}. This task is called time-series causal discovery, which serves as a fundamental tool in machine learning by enabling causal reasoning of time-series. 

Although proven to be able to efficiently discover causal relationships, most of these methods lack the ability to handle high-dimensional time-series. Actually, many causal discovery algorithms are only tested on datasets with fewer than 20 time-series (i.e. $N\leq 20$) \cite{tankNeuralGrangerCausality2022, khannaEconomyStatisticalRecurrent2020}, while the real time-series datasets often contain dozens or even hundreds of time-series, e.g., gene regulation networks or air quality index. Recently, \citet{chengCUTSNeuralCausal2023} proposed CUTS, an iterative approach to jointly perform causal graph learning and missing data imputation for irregular temporal data. Although CUTS is proposed to boost causal discovery with data imputation and the other way around, the data prediction module is composed of component-wise LSTMs and MLPs with redundant structures and parameters, hampering the scalability when encountering high-dimensional datasets. Moreover, the causal graph in CUTS can be too large to learn with high accuracy.

To overcome these issues, we propose \textbf{CUTS+}, an extension of CUTS with scalability to high-dimensional time-series, via proposing two specially designed techniques: coarse-to-fine causal discovery (C2FD) and message-passing graph neural network (MPGNN) for data prediction. Our contributions include:
\begin{itemize}
    \item We propose CUTS+, upgrading CUTS \cite{chengCUTSNeuralCausal2023} to largely increase the scalability towards high-dimensional time-series. Specifically, we leverage two novel techniques, i.e., Coarse-to-fine-discovery (C2FD), a simple yet efficient technique to facilitate scalable causal graph optimization, and message-passing-based graph neural network (MPGNN) to remove structural redundancy in CUTS+.
    \item With extensive experiments, we show that CUTS+ largely increases causal discovery performance and decreases time cost, especially on high-dimensional datasets, with either multiple types of irregular sampling or no missing values. 
\end{itemize}

\section{Related Works}

\textbf{Causal Structural Learning / Causal Discovery.~~~~}
Existing Causal Structural Learning (or Causal Discovery) approaches can be categorized into five classes. (i) \textit{Constraint-based approaches}, such as PC \cite{spirtesAlgorithmFastRecovery1991}, FCI \cite{spirtesCausationPredictionSearch2000}, and PCMCI \cite{rungeDetectingQuantifyingCausal2019, rungeDiscoveringContemporaneousLagged2020, gerhardusHighrecallCausalDiscovery2020}, build causal graphs by conditional independence tests. (ii) \textit{Score-based learning algorithms} which include penalized Neural Ordinary Differential Equations and acyclicity constraint\cite{bellotNeuralGraphicalModelling2022} \cite{pamfilDYNOTEARSStructureLearning2020}. (iii) \textit{Convergent Cross Mapping (CCM)} proposed by \citet{sugiharaDetectingCausalityComplex2012} that reconstructs nonlinear state space for nonseparable weakly connected dynamic systems. This approach is later extended to situations of synchrony, confounding, or sporadic time series \citep{yeDistinguishingTimedelayedCausal2015, benkoCompleteInferenceCausal2020, brouwerLatentConvergentCross2021}. (iv) Approaches based on \textit{Additive Noise Model (ANM)} that infer causal graph based on additive noise assumption \citep{shimizuLinearNonGaussianAcyclic2006, hoyerNonlinearCausalDiscovery2008}. ANM is extended by \citet{hoyerNonlinearCausalDiscovery2008} to nonlinear models with almost any nonlinearities. (v) \textit{Granger-causality-based} approaches. Granger causality is initially introduced by \citet{grangerInvestigatingCausalRelations1969} who proposed to analyze the temporal causal relationships by testing the help of a time-series on predicting another time-series. Recently, Deep Neural Networks (NNs) are widely applied to infer nonlinear Granger causality since the central idea of Granger Causality is highly compatible with NNs. Researchers have successfully use Recurrent Neural Networks (RNNs) or other NNs for time series analysis to discover causal graphs \cite{wuGrangerCausalInference2022, tankNeuralGrangerCausality2022, khannaEconomyStatisticalRecurrent2020, loweAmortizedCausalDiscovery2022, chengCUTSNeuralCausal2023}. This work also incorporates a deep neural network to discover Granger Causality.

\textbf{Scalable / High-dimensional Causal Discovery.~~~~} Scalability can be a serious problem when applying causal discovery algorithms to real data. With hundreds of time-series (or hundreds of static nodes), the potential possibility for causal relations grows exponentially. Existing approaches may fail because they involve either massive conditional independence tests \cite{rungeDetectingQuantifyingCausal2019}, too many variables to be conditioned on \cite{hongEfficientAlgorithmLargescale2017}, or large quantities of parameters to be optimized \cite{tankNeuralGrangerCausality2022, chengCUTSNeuralCausal2023}. To solve this problem, scalable or high-dimensional causal discovery approaches are proposed. In static settings, \citet{hongEfficientAlgorithmLargescale2017} and \citet{morales-alvarezSimultaneousMissingValue2022} propose to boost scalability via 
divide-and-conquer technique, \citet{lopezLargeScaleDifferentiableCausal2022a} limit the search space to low-rank factor graphs, \citet{cundyBCDNetsScalable2021} instead leverages variational framework. In time-series settings like ours, the scalability issue is less explored. The most related work to ours is \citet{xuScalableCausalGraph2019}'s which also uses Granger causality and simplifies the high-dimensional adjacency matrix with low-rank approximation. However, the low-rank assumption may not be satisfied in real scenarios.  Our CUTS+ is an extension of Granger-causality-based approaches by alleviating the scalability issue without low-rank approximation. 

\section{Background}
\subsection{Time Series and Granger Causality}

We inherit the notation in \cite{chengCUTSNeuralCausal2023} and denote a uniformly sampled observation of a dynamic system as $\mathbf{X} = \{\mathbf{x}_{i,1:T} \}_{i=1}^{N}$ , where $x_{i,t}$  represents the $i$th time-series sampled at time point $t$, and $t\in \left\{ 1, ..., T \right\}$, $i\in \left\{ 1, ..., N \right\}$, with $T$ and $N$ being the length and number of the time-series. Each sampled variable $x_{i,t}$ is assumed to be generated by the following Structural Causal Model (SCM) with additive noise:
\begin{equation}
\label{gen}
    x_{i,t}=f_i(\mathbf{x}_{1, t-\tau:t-1}, \mathbf{x}_{2, t-\tau:t-1}, ..., \mathbf{x}_{N, t-\tau:t-1}) + e_{i, t}
\end{equation}
in which $\tau$ denotes the maximal time lag and $i=1,2,..., N$. Our CUTS+ can also handle irregular time-series by jointly performing imputation and causal discovery. So to model the irregular time-series, a bi-value observation mask $o_{t,i}$ is used to label the missing entries, i.e., the observed point equals the generated $x_{i,t}$ when $o_{t,i}$ equals to 1. In this paper, we adopt the protocols of previous works \cite{yiSTMVLFillingMissing2016, ciniFillingApMultivariate2022} and consider two types of data missing that often occur in practical observations:

\begin{itemize}[leftmargin=*]
    \item \textit{Random Missing (RM). } The data entries in the observations are missing with a certain probability $p$, here in our experiments the missing probability follows Bernoulli distribution $o_{t,i}\sim Ber(1-p)$.
    \item \textit{Random Block Missing (RBM). } Under a relatively small $p$ for RM, we set a block failure probability $p_{\text{blk}}$ and block length $L_{\text{blk}}\sim \text{Uniform}(L_{\text{min}}, L_{\text{max}})$, i.e. there exist $p_{\text{blk}}\cdot N\cdot T$ missing blocks on average and each with length uniformly distributed in $[L_{\text{min}}, L_{\text{max}}]$.
\end{itemize}

Note that these two types can both be categorized into Missing Complete at Random (MCAR), a most common type of data missing \cite{geffnerDeepEndtoendCausal2022}. In this work, we build on the Granger causality. Actually, Granger causality is not necessarily SCM-based causality, since the latter one often considers acyclicity. Under the assumptions of no unobserved variables and no instantaneous effects, \citet{petersElementsCausalInference2017} shows identifiablility of time-invariant Granger causality \cite{loweAmortizedCausalDiscovery2022, vowelsYaDAGsSurvey2021}. For a dynamic system, time-series $i$ Granger causes time-series $j$ when the past values of time-series $\mathbf{x}_i$ aid in the prediction of the current and future status of time-series $\mathbf{x}_j$. Specifically, we adapt the definition from \cite{tankNeuralGrangerCausality2022} that time-series $i$ is Granger non-causal of $j$ if there exists $\mathbf{x}'_{i,t-\tau:t-1} \neq \mathbf{x}_{i,t-\tau:t-1}$ satisfying
\begin{equation}
\begin{aligned}
    & f_{j}(\mathbf{x}_{1, t-\tau:t-1},... , \mathbf{x}'_{i, t-\tau:t-1}, ..., \mathbf{x}_{N, t-\tau:t-1}) = \\
    & f_{j}(\mathbf{x}_{1, t-\tau:t-1}, ...,  \mathbf{x}_{i, t-\tau:t-1}, ..., \mathbf{x}_{N, t-\tau:t-1})
\end{aligned}
\end{equation}
i.e., the past data points of time-series $i$ influence the prediction of $x_{j, t}$. For simplicity, we use $\mathbf{x}_i$ to denote $\mathbf{x}_{i, t-\tau:t-1}$ in the following. The discovered pair-wise Granger causal relationship is a directed graph, which is then represented with an adjacency matrix $\mathbf{A}=\{a_{ij}\}_{i,j=1}^{N}$, where $a_{ij} = 1$ denotes time-series $i$ Granger causes $j$ and $a_{ij} = 0$ means otherwise. The idea of Granger causality is highly compatible with the basic idea of neural networks (NNs) since NNs can serve as powerful predictors. In previous works, \cite{chengCUTSNeuralCausal2023} prove that causal graph discovery in their CUTS converges to the true graphs under the Additive Noise Model and Universal Approximation Theorem, which again validates the successful combination of Granger causality and NNs. 

\subsection{Difficulties with High Dimensional Time-series}
\label{difficulty}

In real scenarios, it is common to face hundreds of long time-series with complex causal graphs. We now proceed to show difficulties when CUTS or other causal discovery algorithms handle such data. 

\textbf{Large Adjacency Matrix.~~~~} The pairwise GC relationship, denoted as matrix $\mathbf{A}_{N\times N}$, can become very large as $N$ increases. Prior works mainly focus on the settings when $N \leq 20$ \cite{tankNeuralGrangerCausality2022, khannaEconomyStatisticalRecurrent2020, chengCUTSNeuralCausal2023}, and we show in the experiments that their performances degrade greatly when $N>20$. Alternatively, \citet{xuScalableCausalGraph2019} addresses the scalability issue at increasing data dimension via conducting low-rank approximation to the adjacency matrix, but the strong low-rank assumption of $\mathbf{A}_{N\times N}$ does not hold in many scenarios. 

\textbf{Redundancy of cMLP / cLSTM.~~~~}  To uncover the black-box of NNs, \citet{chengCUTSNeuralCausal2023, tankNeuralGrangerCausality2022} disentangle causal effects from causal parents to individual output series. As a result, one must use $N$ separate MLPs / LSTMs to ensure the disentanglement. This is called component-wise MLP / LSTM (cMLP / cLSTM) and frequently used when discovering Granger causality \cite{khannaEconomyStatisticalRecurrent2020}. In the following We formalize component-wise neural networks as ``causally disentangled neural networks''.

\begin{definition}
\label{cdnn}
    Let $\mathbf{x} \in \mathbf{\mathcal{X}} \subset \mathbb{R}^{n}, \mathbf{A} \in \mathbf{\mathcal{A}} \subset \left\{0,1\right\}^{n\times n}$ and $\mathbf{y} \in \mathbf{\mathcal{Y}} \subset \mathbb{R}^n$ be the input and output spaces. We say a neural network $\mathbf{f}_\Theta: \left<\mathbf{\mathcal{X}}, \mathbf{\mathcal{A}} \right> \rightarrow \mathbf{\mathcal{Y}}$ is a \textbf{causally disentangled neural network (CDNN)} if it has the form 
    \begin{equation}
        \mathbf{f}_\Theta(\mathbf{x}, \mathbf{A}) = \left[ f_{\Theta_1}(\mathbf{x} \odot \mathbf{a}_{:,1}), ..., f_{\Theta_n}(\mathbf{x} \odot \mathbf{a}_{:,n}) \right]^T.
    \end{equation}
    Here $\mathbf{a}_{:,j}$ is the column vector of input causal adjacency matrix $\mathbf{A}$; $f_{\phi_j}: \mathbf{\mathcal{X}}_j \rightarrow \mathbf{\mathcal{Y}}_j$, with $\mathbf{\mathcal{X}}_j \subset \mathbb{R}^{n}$ and $\mathbf{\mathcal{Y}}_j \subset \mathbb{R}$; the operator $\odot$ denotes the Hadamard product.
\end{definition}

Here function $f_{\Theta_j}(\cdot)$ represents the neural network function used to approximate $f_j(\cdot)$ in Equation (\ref{gen}). The input to CDNN can also be $\mathbf{X}$ with time dimension instead of $\mathbf{x}$, then $\odot$ is defined as $f_{\phi_j}(\mathbf{X} \odot \mathbf{a}_{:,j}) \overset{\Delta}{=} f_{\phi_j}\left(\left\{ \mathbf{x}_1 \cdot a_{1j}, ..., \mathbf{x}_N \cdot a_{Nj} \right\}\right)$. Under this definition, \citet{chengCUTSNeuralCausal2023} proved that when approximating $f_j$ with $f_{\phi_j}$ (along with other assumptions), the discovered causal adjacency matrix will converge to the true Granger causal matrix.
Although being a CDNN, cMLP / cLSTM consists of $N$ separate networks and is highly redundant, because the shared dynamics among different time-series are modeled $N$ times. 
This redundancy not only slows down the learning process but also degrades causal discovery accuracy. Therefore, the model does not scale well to high-dimension time-series. In the following section, we introduce two techniques to alleviate the scalability issue.

\begin{table}
\label{acronyms}
\centering
\footnotesize
\begin{tabular}{ll}
\toprule[.4pt]
CPG     & Causal Probability Graph $M$              \\
GCPG    & Group CPG $Q$                             \\
BCG     & Binary Causal Graph    $S$                \\
MPNN    & Message-Passing NN                        \\
MPGNN   & Message-Passing GNN                       \\
C2FD    & Coarse-to-fine Causal Discovery           \\
\bottomrule[.4pt]
\end{tabular}
\caption{List of abbreviations.}
\end{table}

\section{CUTS+}

In this work, we implement the causal graph as Causal Probability Graphs (CPGs) $\mathbf{M}$ in which the element $m_{ij}$ represents the probability of time-series $i$ Granger causing $j$, i.e., $m_{ij}=P(x_{i} \rightarrow x_{j})$. If $\tilde{m}_{ij}$ in the discovered graph $\mathbf{\tilde{\mathbf{M}}}$ is penalized to zero (or below some certain threshold), we can deduce that time-series $i$ does not Granger cause $j$. 

\begin{figure*}[t]
    \begin{center}
        \includegraphics[width=\textwidth]{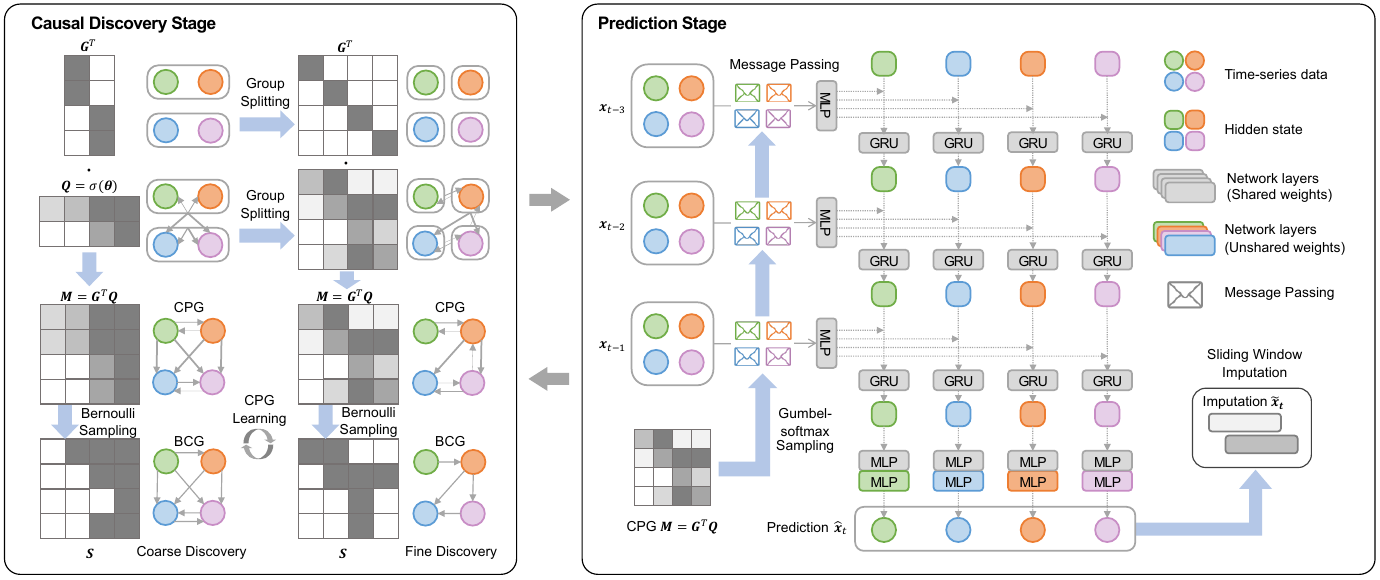}
    \end{center}
    
    \caption{The architecture of CUTS+ with two alternating stages, both boosted for high-dimensional causal discovery. The \textbf{\textit{Causal Discovery Stage}} is equipped with Coarse-to-fine Causal Discovery (C2FD) while the \textbf{\textit{Prediction Stage}} is with Message-passing-based Graph Neural Network (MPGNN).}

    \label{fig:cutsp}
\end{figure*}

Similar to CUTS \cite{chengCUTSNeuralCausal2023}, we also adapt a two-stage training strategy, and iteratively perform the \textbf{\textit{Causal Discovery Stage}} and \textbf{\textit{Prediction Stage}}---the former builds a causal probability matrix with available time-series under sparse penalty, while the latter one fits the complex distribution of high-dimensional time-series and fills the missing entries. However, both stages are of totally new designs to overcome the difficulties when encountering high-dimensional time-series, as illustrated in Figure~\ref{fig:cutsp}. Specifically, we propose to use the coarse-to-fine discovery (C2FD) technique in the Causal Discovery Stage 
and message-passing graph neural network (MPGNN) in the Prediction Stage, 
which are detailed in the following subsections.


\subsection{Coarse-to-fine Causal Discovery}

To address the problem of \textbf{large adjacency matrix} discussed before, we propose Coarse-to-Fine causal Discovery (C2FD). Specifically, we split the time-series into several groups, i.e., time-series group $\mathbf{X}_{\mathcal{G}_k} = \left\{ \mathbf{x}_i \right\}_{i\in \mathcal{G}_k}$ where $\mathcal{G}_k$ is the set of the indices within the $k$th group, and $k \in \left[ 1, N_g\right]$ with $N_g$ being the group number. Each time-series is and can only be allocated to one group, i.e., $\forall k\neq l  \in \left[ 1, N_g\right], \mathcal{G}_k\cap \mathcal{G}_l = \phi$. The grouping is implemented with matrix decomposition of $\mathbf{\tilde{\mathbf{M}}}$: 
\begin{equation}
    \mathbf{\tilde{\mathbf{M}}} = \mathbf{G}^T \mathbf{Q}
\end{equation}
where $\mathbf{G} \in \mathbb{R}^{N_g \times N}$ is composed of entries 
$g_{ki}=\begin{cases}
    0, & \text{if } i \notin \mathcal{G}_k\\
    1, & \text{if } i \in \mathcal{G}_k,
\end{cases}$. Since each time-series is and can only be allocated to one group, the sum of each column vector of $\mathbf{G}$ is 1, i.e. $\forall i \in \left[1, N \right], \sum_{k=1}^{N_g} g_{ki} = 1$. We define the Granger causality from group $\mathbf{X}_{\mathcal{G}_k}$ as follows
\begin{definition}
Group $\mathbf{X}_{\mathcal{G}_k}$ is Granger non-causal of $\mathbf{x}_{j}$ if there exists  $\mathbf{X}'_{\mathcal{G}_k} \neq \mathbf{X}_{\mathcal{G}_k}$, 
\begin{equation}
    f_{j}\left( \left\{ \mathbf{X}'_{\mathcal{G}_k}, \mathbf{X} \backslash \mathbf{X}_{\mathcal{G}_k}\}\right) = f_{j} \left( \{ \mathbf{X}_{\mathcal{G}_k}, \mathbf{X} \backslash \mathbf{X}_{\mathcal{G}_k}\right\} \right)
\label{groupgc}
\end{equation}
i.e., group $x_{\mathcal{G}_k}$ influence the prediction of $x_{j,t}$. Here we define $\mathbf{X} \backslash \mathbf{X}_{\mathcal{G}_k} \overset{\Delta}{=} \left\{ \mathbf{x}_i \right\}_{i \notin \mathcal{G}_k}$.
\end{definition} 
Then $\mathbf{Q} \in \mathbb{R}^{N_g\times N}$ is the Group Causal Probability Graph (GCPG) with $q_{ij}=P(\mathbf{X}_{\mathcal{G}_i} \rightarrow \mathbf{x}_{j})$. 

\textbf{Initial Allocation.~~~~} Before training, we initiate $\mathbf{G}$ with a relatively small $N_g$ and consequently obtain a ``coarse'' grouping. Specifically, the time series are allocated into a group following
\begin{equation}
    |\mathcal{G}_i|=\begin{cases}
    \lfloor N / N_g \rfloor, & \text{if } i \in \left[ 1, N_g - 1\right]\\
    N  - \lfloor N / N_g \rfloor \cdot (N_g - 1), & \text{if } i = N_g.
\end{cases}
\end{equation}

\textbf{Group Splitting.~~~~} During training, we periodically split each group into two groups, and then $N_g$ is doubled every 20 epochs until every group contains only one time-series. Correspondingly, when doubling group numbers, GCPG element $q_{ij}$ is allocated to 2 elements ${q}_{i_1, j}$ and ${q}_{i_2, j}$ in the new GCPG, as initial guesses. Here we assume a group Granger cause $\mathbf{x}_{j}$ if at least one sub-group Granger cause $\mathbf{x}_{j}$, then $q_{ij} =P(\mathbf{X}_{\mathcal{G}_i} \rightarrow \mathbf{x}_{j})=1 - P(\mathbf{X}_{\mathcal{G}_{i_1}} \nrightarrow \mathbf{x}_{j})P(\mathbf{X}_{\mathcal{G}_{i_2}} \nrightarrow \mathbf{x}_{j})$, where $\nrightarrow$ denotes not Granger cause. As initial guesses we assume $P(\mathbf{X}_{\mathcal{G}_{i_1}} \rightarrow \mathbf{x}_{j}) = P(\mathbf{X}_{\mathcal{G}_{i_2}} \rightarrow \mathbf{x}_{j})$, then $q_{i_1, j}$ and $q_{i_2, j}$ are calculated as
\begin{equation}
    q_{i_1, j} = q_{i_2, j} = 1 - \sqrt{1 - q_{ij}}.
\end{equation}

\textbf{Convergence of C2FD.~~~~} We prove the convergence of C2FD with CDNN as predictor, with a similar manner to \cite{chengCUTSNeuralCausal2023}. Due to space limit, we place the detailed assumptions, theorem, and proof in Section A of the supplementary material.

The idea of coarse-to-fine is quite common in the field of machine learning \cite{fleuretCoarsetoFineFaceDetection2001, sarlinCoarseFineRobust2019}. However, to our best knowledge, we are the first to introduce the idea into neural-network-based Granger causality. Actually, the advantage of introducing C2FD is two-fold. Firstly, the parameter number to learn is greatly decreased in the initial stages. In the initial stages when $N_g \ll N$, only $N\cdot N_g$ instead of $N^2$ parameters are required to be optimized. Secondly, the learning results with smaller $N_g$ serve as initial guess for learning with larger $N_g$. When $|\mathcal{G}_k| > 1$, GCPG element $q_{kj}$ increases towards $1$ if at least one member Granger cause time-series $j$. Then the whole group is used to perform data prediction with higher accuracy. After doubling $N_g$, the optimizer further locates the sub-group that actually Granger causes $j$. The empirical advantage of C2FD is validated in experiments section.

\subsection{Message-passing-based Graph Neural Network}

To satisfy the definition of causally disentangled neural network (CDNN) while preventing using highly redundant cMLP / cLSTM, we leverage the Message-Passing Neural Network (MPNN \cite{gilmerNeuralMessagePassing2017}) for data prediction encoder. To learn the dynamics of the high-dimensional time-series, we alter the gated recurrent units (GRUs, \cite{choLearningPhraseRepresentations2014}) by adding message-passing layers. 
Firstly, we formulate single-layer MPNN as 
\begin{equation}
    \text{MPNN}_{\mathbf{\nu}}(\mathbf{z}; \mathbf{s}) = \text{MLP}_{\mathbf{\nu}}\left(\mathbf{z} \odot \mathbf{s} \right) = \mathbf{z}'
\end{equation}
where $\mathbf{h}'$ is the output of MPNN in the last layer, $\text{MLP}_1(\cdot), \text{MLP}_2(\cdot)$ is multi-layer perceptrons (MLPs), $\odot$ denotes the Hadamard product, and $\mathbf{s} \in \left\{0,1\right\}^N$ is the binary causal vector, i.e., a column in the sampled Binary Causal Graph (BCG, described in Section \ref{arch}) where $s_i=1$ denotes $z_i$ Granger cause the prediction. Similar to  \cite{ciniFillingApMultivariate2022}, we add MPNN to GRU units, which serves as a layer in MPGNN:
\begin{align}
\mathbf{r}_t^j &= \sigma \left(\text{MPNN}_{\mathbf{\nu}_l^r}\left( {\mathbf{x}}_{:,t} ; \mathbf{s}_{:,j} \right)  \right)            \\
\mathbf{u}_t^j &= \sigma \left(\text{MPNN}_{\mathbf{\nu}_l^u}\left( {\mathbf{x}}_{:,t} ; \mathbf{s}_{:,j} \right)  \right)            \\
\mathbf{c}_t^j &= \text{tanh} \left(\text{MPNN}_{\mathbf{\nu}_l^c}\left( {\mathbf{x}}_{:,t} ; \mathbf{s}_{:,j} \right) \right)       \\
\mathbf{h}_t^j &= \mathbf{u}_t^i \odot \mathbf{h}_{t-1}^i + (1 - \mathbf{u}_t^i) \odot \mathbf{c}_t^i,
\end{align}
 with $\sigma(\cdot)$ being the sigmoid function. Different from standard GRU units, each gate is computed using only the input vector, which decreases parameter numbers. In the following we represent $l$ MPGNN layers with $\text{MPGNN}_{\mathbf{\nu}} \left({\mathbf{x}}_{:,t}, \mathbf{h}_{0}^j; \mathbf{s}_{:,j} \right)$, where $\mathbf{\nu} = \{\mathbf{\nu}^r_1, \mathbf{\nu}^u_1, \mathbf{\nu}^c_1, ..., \mathbf{\nu}^r_l, \mathbf{\nu}^u_l, \mathbf{\nu}^c_l\}$ (parameters of MPNNs in all layers). Note that we share $\mathbf{\nu}$ for each $j$, which is the key design contributing to high scalability. 

\textbf{Scalability of MPGNN Encoder.~~~~} The number of parameters that need to be optimized in the MPGNN encoder can be calculated as $l \left( |\mathbf{\nu}^r| + |\mathbf{\nu}^u| + |\mathbf{\nu}^c|\right)$, where $l$ is the number of MPGNN layers. Comparing CUTS+ with component-wise GRU, whose parameter number is $Nl \left( |\mathbf{\nu}^r| + |\mathbf{\nu}^u| + |\mathbf{\nu}^c|\right)$ (or cMLP / cLSTM \cite{tankNeuralGrangerCausality2022} which is also $O(Nl)$), MPGNN achieves high scalability by significantly reducing the number of optimization parameters in the encoder. Moreover, since the component-wise network-based prediction model is usually overparameterized and thus prone to overfitting \cite{khannaEconomyStatisticalRecurrent2020}, our design also helps to mitigate overfitting.

\textbf{Decoder.~~~~} After encoding with MPGNN, the prediction result $\hat{x}_{j,t}$ is retrieved with a two-part decoder, i.e., 
\begin{equation}
    \hat{x}_{j,t} = \text{Linear}_{\psi_j^2}\left(\text{MLP}_{\psi^1}\left(\mathbf{h}_{t-1}^{j} \right)\right)
\end{equation}
where $\text{Linear}_{\psi_j^2}(\cdot)$ denotes a single linear layer with unshared weights. To capture the heterogeneity among time-series while removing structural redundancy as much as possible, here we share the weights of the first MLP part ($\psi$) and do not share the weight of the second single-layered part (distinct $\psi_j$ for each target time-series $j$). 

\subsection{Overall Architecture}
\label{arch}

\textbf{Bernoulli Sampling of CPG.~~~~} Our CUTS+ represents causal relationships with CPG $\tilde{\mathbf{M}}$. To ensure elements of $\tilde{\mathbf{M}}$ are within range $\left[0, 1\right]$, we set $\mathbf{Q} = \sigma(\mathbf{\Theta})$ where $\mathbf{\Theta}$ is the parameter to learn. During \textbf{\textit{Causal Discovery Stage}} $\mathbf{\Theta}$ is optimized using Gumbel-Softmax estimator \cite{jangCategoricalReparameterizationGumbelsoftmax2016}, i.e., 
\begin{equation}
\label{gumbel}
    s_{ij} = \frac{e^{{(\log(m_{ij})+g)}/{\tau}}}{e^{{(\log(m_{ij})+g)}/{\tau}} + e^{{(\log(1-m_{ij})+g)}/{\tau}}}
\end{equation} 
Where $g=-\log(-\log(u)), u\sim \text{Uniform}(0,1)$. 
We use a large $\tau$ in the initial stages then decrease to a small value. This estimator has a relatively low variance, mimic Bernoulli distribution when $\tau$ is small, and more importantly, enables continuous optimization of $\mathbf{\Theta}$. 

During \textbf{\textit{Prediction Stage}}, CPG $\tilde{\mathbf{M}}$ is sampled to binary causal graph (BCG) $\mathbf{S}$ where $s_{ij}\sim \text{Ber}(m_{ij})$. Finally BCG columns $\mathbf{s}_{:,j}$ is used as adjacency matrix in MPNNs. 

\textbf{Loss Functions.~~~~} CUTS+ iterates between \textbf{\textit{Causal Discovery Stage}} and \textbf{\textit{Prediction Stage}}. During the former stage, only the CPG $\tilde{\mathbf{M}} = \sigma(\mathbf{\Theta})$ will be optimized, so the optimization problem is

\begin{equation}
\begin{aligned}
    \max_{\mathbf{\Theta}} \mathcal{L}_{\text{graph}} = \max_{\mathbf{\Theta}} & \frac{\sum_{j=1}^{N}\sum_{t=1}^{T} \left( \hat{x}_{j,t} - x_{j,t}  \right)^2\cdot o_{j,t}}{\sum_{j=1}^{N}\sum_{t=1}^{T} o_{j,t}} \\
    & + \lambda \left\| \sigma(\mathbf{\Theta}) \right\|_1.
\end{aligned}
\end{equation}
In the latter stage, we only optimize the network parameters:
\begin{equation}
    \max_{\Phi} \mathcal{L}_{\text{data}} = \max_{\Phi} \frac{\sum_{j=1}^{N}\sum_{t=1}^{T} \left( \hat{x}_{j,t} - x_{j,t}  \right)^2\cdot o_{j,t}}{\sum_{j=1}^{N}\sum_{t=1}^{T} o_{j,t}}.
\end{equation}
where $\Phi = \left\{ \nu, \psi_1, \{\psi_2^j\}_{j=1}^N \right\}$ is all network parameters in MPGNN encoder and decoder.

\textbf{Satisfaction of CDNN.~~~~} CDNN in Definition \ref{cdnn} is satisfied when a column of the matrix $\mathbf{A}$ only affects the corresponding component of the prediction $\mathbf{f}_{\Phi}(x, \mathbf{A})$. If we combine the prediction module $\text{MPGNN}_{\mathbf{\nu}} \left(\tilde{\mathbf{X}}, \mathbf{h}_{0}^j; \mathbf{s}_{:,j} \right)$ with $\text{Linear}_{\psi_j^2}\left(\text{MLP}_{\psi^1}\left( \cdot \right)\right)$, we get 
\begin{equation}
\begin{aligned}
    &\hat{x}_{:,t} = \mathbf{f}_{\Phi}(\mathbf{X}, \mathbf{S}) =\\
    & \left[...,  \text{Linear}_{\psi_j^2}\left(\text{MLP}_{\psi^1}\left( \text{MPGNN}_{\mathbf{\nu}} \left(\tilde{\mathbf{X}}, \mathbf{h}_{0}^j; \mathbf{s}_{:,j} \right) \right) \right), ...\right]^T
\end{aligned}
\end{equation}
where $\mathbf{h}_{0}^j$ is the initial value of GRU hidden states and irrelevant to $\mathbf{x}$. Therefore, our CUTS is a CDNN, and according to Theorem 1 in supplementary Section A, the correct causal graph can be recovered. 

\textbf{Handling Irregular Time-series with Imputation.~~~~} In this work, we handle irregular time-series by performing concurrent data imputation during \textbf{\textit{Prediction Stage}}. Our data imputation is performed with sliding windows, 
where the missing entries in one time window are filled with the prediction $\hat{x}_{j,t}$ from the last windows. Due to page limits, we place the details for sliding window imputation in supplementary Section C.3.

\begin{table*}[t]
\small
\centering
\begin{tabular}{cc|cc|cc|c}
\\ \toprule[.7pt]
\multirow{2}{*}{\textbf{Method}} & \multirow{2}{*}{\textbf{Imputation}}   & \multicolumn{2}{c|}{\textbf{VAR with RM ($N=128$)}}      & \multicolumn{2}{c|}{\textbf{VAR with RBM ($N=128$)}}      & \textbf{VAR ($N=128$)}      \\
                                  &              &~~ $p=0.3$~~                      & ~~$p=0.6$~~                     & ~~$p_{\text{blk}}=0.15\%$~~                 & $~~p_{\text{blk}}=0.3\%~~$            & No missing    \\ \midrule[.4pt]
\multirow{2}{*}{NGC}              & ZOH                             & 0.8234 {$\pm$ 0.0082}          & 0.7419 {$\pm$ 0.0056}         & 0.8638 {$\pm$ 0.0165}           & 0.8357 {$\pm$ 0.0161}      & \multirow{2}{*}{0.9168 {$\pm$ 0.0087}}     \\
                                  & TimesNet                        & 0.7900 {$\pm$ 0.0111}          & 0.6560 {$\pm$ 0.0112}         & 0.8519 {$\pm$ 0.0056}           & 0.8292 {$\pm$ 0.0083}         \\
\multirow{2}{*}{eSRU}             & ZOH                             & 0.6627 {$\pm$ 0.0071}          & 0.6711 {$\pm$ 0.0097}         & 0.6606 {$\pm$ 0.0152}           & 0.6457 {$\pm$ 0.0060}      & \multirow{2}{*}{0.6860 {$\pm$ 0.0144}}    \\
                                  & TimesNet                        & 0.6175 {$\pm$ 0.0149}          & 0.5671 {$\pm$ 0.0156}         & 0.6643 {$\pm$ 0.0097}           & 0.6488 {$\pm$ 0.0096}          \\
\multirow{2}{*}{SCGL}             & ZOH                             & 0.6627 {$\pm$ 0.0071}          & 0.6711 {$\pm$ 0.0097}         & 0.6606 {$\pm$ 0.0152}           & 0.6457 {$\pm$ 0.0060}      & \multirow{2}{*}{0.6519 {$\pm$ 0.0078}}    \\
                                  & TimesNet                        & 0.6536 {$\pm$ 0.0180}          & 0.6542 {$\pm$ 0.0048}         & 0.6558 {$\pm$ 0.0118}           & 0.6631 {$\pm$ 0.0138}         \\
\multicolumn{2}{c|}{NGM}                                            & 0.5815 {$\pm$ 0.0494}          & 0.5016 {$\pm$ 0.0010}         & 0.5918 {$\pm$ 0.0700}           & 0.5003 {$\pm$ 0.0004}      & 0.6626 {$\pm$ 0.0052}   \\ \midrule[.4pt]
\multicolumn{2}{c|}{CUTS}                                           & 0.9434 {$\pm$ 0.0123}          & 0.8814 {$\pm$ 0.0151}         & 0.9579 {$\pm$ 0.0085}           & 0.9505 {$\pm$ 0.0091}      & 0.9626 {$\pm$ 0.0057}   \\
\multicolumn{2}{c|}{CUTS with C2FD}                                    & 0.9594 {$\pm$ 0.0094}          & 0.8752 {$\pm$ 0.0183}         & 0.9742 {$\pm$ 0.0061}           & 0.9651 {$\pm$ 0.0072}      & 0.9875 {$\pm$ 0.0024}   \\
\multicolumn{2}{c|}{\textbf{CUTS+}}                                 & \textbf{0.9907 {$\pm$ 0.0008}} & \textbf{0.9569 {$\pm$ 0.0051}}& \textbf{0.9939 {$\pm$ 0.0018}}  & \textbf{0.9912 {$\pm$ 0.0025}} & \textbf{0.9972 {$\pm$ 0.0005}}  \\ \midrule[.7pt]
\multirow{2}{*}{\textbf{Method}} & \multirow{2}{*}{\textbf{Imputation}}   & \multicolumn{2}{c|}{\textbf{Lorenz-96 with RM ($N=256$)}} & \multicolumn{2}{c|}{\textbf{Lorenz-96 with RBM ($N=256$)}} & \textbf{Lorenz-96 ($N=256$)}      \\
                                  &                                 & $p=0.3$                      & $p=0.6$                     & $p_{\text{blk}}=0.15\%$                 & $p_{\text{blk}}=0.3\%$             & No missing       \\ \midrule[.4pt]
\multirow{2}{*}{NGC}              & ZOH                             & 0.9755 {$\pm$ 0.0092}          & 0.8469 {$\pm$ 0.0331}         & 0.9893 {$\pm$ 0.0022}         & 0.9760 {$\pm$ 0.0042}          & \multirow{2}{*}{0.9937 {$\pm$ 0.0014}}     \\
                                  & TimesNet                        & 0.9415 {$\pm$ 0.0183}          & 0.5000 {$\pm$ 0.0000}         & 0.9685 {$\pm$ 0.0070}         & 0.7965 {$\pm$ 0.0442}        \\
\multirow{2}{*}{eSRU}             & ZOH                             & 0.9735 {$\pm$ 0.0019}          & 0.8972 {$\pm$ 0.0046}         & 0.9821 {$\pm$ 0.0019}         & 0.9728 {$\pm$ 0.0016}          & \multirow{2}{*}{0.9908 {$\pm$ 0.0005}}     \\
                                  & TimesNet                        & 0.9618 {$\pm$ 0.0044}          & 0.8742 {$\pm$ 0.0047}         & 0.9794 {$\pm$ 0.0042}         & 0.9762 {$\pm$ 0.0033}        \\
\multirow{2}{*}{SCGL}             & ZOH                             & 0.6191 {$\pm$ 0.0090}          & 0.5182 {$\pm$ 0.0109}         & 0.6308 {$\pm$ 0.0061}         & 0.6195 {$\pm$ 0.0069}          & \multirow{2}{*}{0.6620 {$\pm$ 0.0083}}     \\
                                  & TimesNet                        & 0.6210 {$\pm$ 0.0032}          & 0.5280 {$\pm$ 0.0060}         & 0.6312 {$\pm$ 0.0072}         & 0.6054 {$\pm$ 0.0034}        \\
\multicolumn{2}{c|}{NGM}                                            & 0.9620 {$\pm$ 0.0072}          & 0.6125 {$\pm$ 0.0372}         & 0.9831 {$\pm$ 0.0031}         & 0.9866 {$\pm$ 0.0006}          & 0.9907 {$\pm$ 0.0010}   \\ \midrule[.4pt]
\multicolumn{2}{c|}{CUTS}                                           & 0.9360 {$\pm$ 0.0043}          & 0.8668 {$\pm$ 0.0043}         & 0.9430 {$\pm$ 0.0030}         & 0.9330 {$\pm$ 0.0053}          & 0.9571 {$\pm$ 0.0027}   \\
\multicolumn{2}{c|}{CUTS with C2FD}                                    & 0.9790 {$\pm$ 0.0016}          & 0.9069 {$\pm$ 0.0036}         & 0.9874 {$\pm$ 0.0009}         & 0.9834 {$\pm$ 0.0015}         & \textbf{0.9992 {$\pm$ 0.0000}}   \\
\multicolumn{2}{c|}{\textbf{CUTS+}}                                 & \textbf{0.9984 {$\pm$ 0.0002}} & \textbf{0.9916 {$\pm$ 0.0016}}& \textbf{0.9989 {$\pm$ 0.0003}}& \textbf{0.9985 {$\pm$ 0.0002}}          & \textbf{0.9992 {$\pm$ 0.0002}}   \\ \bottomrule[.7pt]
\end{tabular}

\caption{Performance comparison of CUTS+ with NGC, eSRU, NGM, SCGL, and CUTS combined with ZOH and TimesNet for imputation. We do not perform comparison experiments with PCMCI and LCCM, because with a large $N$ and $T$, the running times for these two methods are extremely long. Comparisons to them are performed on Dream-3 datasets (Table \ref{Tab: netsimdream}).}
\label{Tab: varlorenz}
\end{table*}

\section{Experiments}\label{sec:Exp}

In this section, we quantitatively evaluate the proposed CUTS+ and comprehensively compare it with state-of-the-art methods to validate our design.

\textbf{Baseline Algorithms.~~~~}   To demonstrate its advantageous performance, we compared CUTS+ with 7 baseline algorithms: (i) Neural Granger Causality (NGC, \cite{tankNeuralGrangerCausality2022}), which utilizes cMLP and cLSTM to infer Granger causal relationships; (ii) economy-SRU (eSRU, \cite{khannaEconomyStatisticalRecurrent2020}), a variant of SRU that is less prone to over-fitting when inferring Granger causality; (iii) PCMCI \cite{rungeDetectingQuantifyingCausal2019}, a non-Granger-causality-based method that uses conditional independence tests; (iv) Latent Convergent Cross Mapping (LCCM, \cite{brouwerLatentConvergentCross2021}), a CCM-based approach that also tackles the irregular time-series problem; (v) Neural Graphical Model (NGM, \cite{bellotNeuralGraphicalModelling2022}), which uses Neural Ordinary Differential Equations (Neural-ODE) to handle irregular time-series data; (vi) Scalable Causal Graph Learning (SCGL, \cite{xuScalableCausalGraph2019}) that address scalable causal discovery problem with low-rank assumption; and (vii) CUTS \cite{chengCUTSNeuralCausal2023}. We evaluated the performance in terms of the area under the ROC curve (AUROC) criterion. For a fair comparison, we search the best hyperparameters for the baseline algorithms on the validation dataset, and test performances on testing sets for 5 random seeds per experiment. For the baseline algorithms that could not handle irregular time-series data, i.e., NGC, PCMCI, SCGL, and eSRU, we imputed the irregular time-series using two algorithms: Zero-order Holder (ZOH, filling with the nearest historical sample, does not introduce future samples), and state-of-the-art imputation algorithm TimesNet \cite{wuTimesNetTemporal2DVariation2023}.

\textbf{Ablation Study Settings.~~~~}  Our main technical contributions are introducing C2FD and MPGNN into causal discovery. To quantitatively validate these two techniques, we add C2FD into the original CUTS \cite{chengCUTSNeuralCausal2023} to get ``CUTS with C2FD''. Consequently, we can measure C2FD's performance gain by comparing ``CUTS'' with ``CUTS with C2FD'' and verify MPGNN by comparing ``CUTS with C2FD'' with ``CUTS+''.

\textbf{Datasets.~~~~}  We assess the performance of the causal discovery approach CUTS+ on three types of datasets: simulated data, quasi-realistic data (i.e., synthesized under physically meaningful causality), and real data. Simulated datasets include linear Vector Autoregressive (VAR) and nonlinear Lorenz-96 models \citep{karimiExtensiveChaosLorenz962010}, quasi-realistic datasets are from Dream-3 \citep{prillRigorousAssessmentSystems2010}, while real datasets include Air Quality datasets from 163 monitor stations across 20 Chinese cities. Irregular observations are generated via Random Missing (RM) and Random Block Missing (RBM). For statistical evaluation of causal discovery algorithms, we average over results on simulations from 5 different random seeds. In the following experiments, we also show the standard derivations.

\subsection{Results on Simulated Datasets}

\textbf{VAR.~~~~}
VAR datasets are simulated with the linear equation $\mathbf{x}_{:,t} = \sum_{\tau=1}^{\tau_{max}} \mathbf{A} \mathbf{x}_{:,t-\tau} + \mathbf{e}_{:,t} $, where the matrix $\mathbf{A}$ is the causal coefficients and $\mathbf{e}_{:,t}\sim \mathcal{N} (\mathbf{0},\sigma\mathbf{I})$. Time-series $i$ Granger cause time-series $j$ if $a_{ij}>0$. The objective of causal discovery is to find the non-zero elements in a causal graph $\mathbf{A}$ with $\tilde{\mathbf{M}}$. We set $\tau_{max}=3$ and time-series length $L=1000$ in this experiment. One can observe in Table \ref{Tab: varlorenz} that CUTS+ beats all other algorithms with a clear margin. Moreover, we perform comparison experiments on VAR datasets with different graph densities in supplementary Section B.1.

\textbf{Lorenz-96.~~~~}
\label{lorenzexp}
Lorenz-96 datasets are simulated according to $\frac{dx_{i,t}}{dt} = -x_{i-1,t}(x_{i-2,t}-x_{i+1,t}) - x_{i,t} + F $, where we set $F=10, L=1000$. In this model each time-series $\mathbf{x}_i$ is affected by historical values of four time-series $\mathbf{x}_{i-2}, \mathbf{x}_{i-1}, \mathbf{x}_{i}, \mathbf{x}_{i+1}$, and each row in the true causal graph $\mathbf{A}$ has four non-zero elements. Table \ref{Tab: varlorenz} shows that CUTS+ is advantageous over other algorithms. 

\textbf{Ablation Study.~~~~} By comparing ``CUTS'', ``CUTS with C2FD'', and ``CUTS+'', we see that both C2FD and MPGNN contribute to the performance gain. C2FD is relatively more helpful on Lorenz-96, and MPGNN helps more on VAR. 

\begin{figure}
\centering
\includegraphics[width=\linewidth]{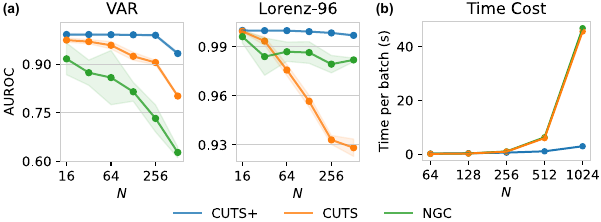}

\caption{Experiments on scalability of the models. (a) Comparison of scalability with CUTS and NGC in terms of AUROC (RM with $p=0.3$). (b) Time cost of CUTS+ comparing with NGC and CUTS+, on $N=64, 128, 256, 512, 1024$.}
\label{Tab: scalab}

\end{figure}

\textbf{Scalability.~~~~} The VAR and Lorenz-96 datasets support setting $N$. To demonstrate CUTS+'s scalability to high dimensional data, we compare CUTS+ with the two best-performing algorithms, i.e., CUTS and NGC (combining ZOH). We set the same average numbers for causal parents in VAR when $N$ changes. Shown in Figure \ref{Tab: scalab}(a), by increasing the time series number $N$ of VAR and Lorenz-96 datasets, we observe that AUROC of CUTS and NGC degrades significantly when $N$ increases on both datasets. On the contrary, CUTS+ beats both algorithms with a clear margin, and the advantages are especially prominent with a large $N$. The performance for CUTS+ only degrades clearly when $N=512$ on VAR datasets. More scalability experiments with multiple types of irregular sampling or no missing values are shown in supplementary Section B.2.

Our advantages over other approaches also exists in terms of computational complexity. Shown in Figure \ref{Tab: scalab}(b) are the time costs for each forward+backward propagation. We compare the network in our CUTS+ with cMLP and cLSTM in NGC and CUTS. The results show that the computational costs are greatly reduced when comparing to cMLP and cLSTM, especially when $N > 256$.

\subsection{Results on Quasi-Realistic Datasets}

\begin{table}[t]
\centering
\small
\begin{tabular}{c|c}
 \toprule[.7pt]
\textbf{Methods}        &~~\textbf{Dream-3} ($N=100$, No missing) ~~                    \\ \midrule[.4pt]
PCMCI                   & 0.5517 {$\pm$ 0.0261}    \\
NGC                     & 0.5579 {$\pm$ 0.0313}    \\
eSRU                    & 0.5587 {$\pm$ 0.0335}    \\
SCGL                    & 0.5273 {$\pm$ 0.0276}    \\
{LCCM}                  & 0.5046 {$\pm$ 0.0318}                     \\
{NGM}                   & 0.5477 {$\pm$ 0.0252}                     \\ \midrule[.4pt]
{CUTS}                  & 0.5915 {$\pm$ 0.0344}                     \\
{CUTS with C2FD}           & 0.6123 {$\pm$ 0.0497}                     \\
\textbf{CUTS+}          & \textbf{0.6374 {$\pm$ 0.0740}}                     \\  \bottomrule[.7pt]
\end{tabular}
\caption{Performance comparison of CUTS+ with PCMCI, NGC, eSRU, NGM, SCGL, LCCM and CUTS on Dream-3 datasets without missing values.}
\label{Tab: netsimdream}
    
\end{table}  


\textbf{Dream-3.~~~~}
Dream-3 \citep{prillRigorousAssessmentSystems2010} is a gene expression and regulation dataset widely used as causal discovery benchmarks \citep{khannaEconomyStatisticalRecurrent2020, tankNeuralGrangerCausality2022}. This dataset contains 5 models, each representing measurements of 100 gene expression levels. The length of each measured trajectory is $T=21$, which is too short to perform RM or RBM, so we only compare with baselines for time-series without data missing on Dream-3. The results are listed in Table \ref{Tab: netsimdream}, which shows that our CUTS+ performs better than the others, proving that our approach can also handle data without missing entries. 

As for the ablation study, we observe that MPGNN and C2FD both contribute clearly, each with performance gain of more than 0.02 in terms of AUROC.

    
%



\subsection{Results on Real Datasets}

\begin{figure}[t]
\includegraphics[width=\linewidth]{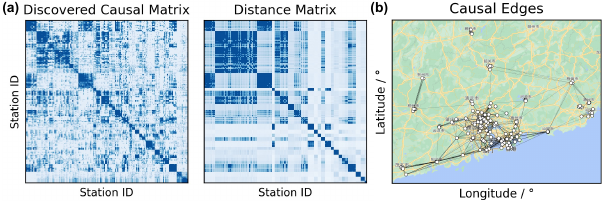}
\caption{Causal discovery result on AQI dataset. (a) Causal discovery results on AQI dataset compared with the distance matrix (which might indicate the true causal graph) (b) Causal discovery results plot overlaid on the map.}
\label{Fig: aqi}

\end{figure}

\textbf{Air Quality (AQI).~~~~} We test our CUTS+ on AQI, a real high-dimensional dataset with $N=163, T=8760$ (detailed description of this dataset is in supplementary Section C.2). We do not have access to the ground-truth causal graph because of the extremely complex atmosphere physics, so quantitative performance evaluation and comparisons with baselines may be less persuasive (shown in supplementary Section B.3). However, we have a prior that the real causal relationships are very closely related to the geometrical distances. Therefore, to verify the causal discovery results of CUTS+, we compare the discovered CPG $\tilde{\mathbf{M}}$ (Figure \ref{Fig: aqi}(a) left) with the distance matrix $\mathbf{D}$ (where its element $d_{ij}\propto 1 / \text{dist}(i, j)$, Figure \ref{Fig: aqi}(a) right). It can be observed that the discovered causal matrix does mimic the distance matrix. We also plotted the CPG edges with $P(i\rightarrow j)>0.5$ on the map (Figure \ref{Fig: aqi}(b)), which shows that most of the causal edges discovered connect the stations not far apart. This indirectly demonstrates the effectiveness of CUTS+ on real high-dimensional data. 

\subsection{Additional Information}

The assumptions, theorems and proof for the convergence of CPG in our CUTS+ are provided in the supplementary Section A. In Section B, we perform other supplementary experiments, including experiments on graph density, scalability, and quantitative comparison on AQI dataset. We provide implementation details for CUTS in Section C, including network structure, hyper-parameters for each experiment, configuration for RM and RBM, and detailed settings for baseline algorithms. Additionally, we list the broader impacts and limitations in Sections D and E.


\section{Conclusions}
We propose CUTS+, a Granger-causality-based causal discovery approach, to handle high-dimensional time-series with irregular sampling. We largely improve the scalability with respect to the data dimension by (a) introducing Coarse-to-fine Discovery (C2FD) to resolve the large CPG problem and (b) designing a Message-passing-based Graph Neural Network (MPGNN) to address the redundant network parameters problem. Comparing to previous approaches, CUTS+ largely increases AUROC and decrease time cost especially when confronted with high dimensional time series.  Our future works include: (i) High-dimensional causal discovery with latent confounder or instantaneous effect. (ii) Explaining neural network with causal models. Our code is available at \texttt{https://github.com/jarrycyx/unn}.

\section{Acknowledgments}
This work is jointly funded by Ministry of Science and Technology of China (Grant No. 2020AAA0108202), National Natural Science Foundation of China (Grant No. 61931012 and 62088102), Beijing Natural Science Foundation (Grant No. Z200021), and Project of Medical Engineering Laboratory of Chinese PLA General Hospital (Grant No. 2022SYSZZKY21).

\appendix

\bibliography{ref}

\begin{thebibliography}{37}
\providecommand{\natexlab}[1]{#1}

\bibitem[{Bellot, Branson, and van~der
  Schaar(2022)}]{bellotNeuralGraphicalModelling2022}
Bellot, A.; Branson, K.; and van~der Schaar, M. 2022.
\newblock Neural Graphical Modelling in Continuous-Time: Consistency Guarantees
  and Algorithms.
\newblock In \emph{International {{Conference}} on {{Learning
  Representations}}}.

\bibitem[{Benk{\H o} et~al.(2020)Benk{\H o}, Zlatniczki, Stippinger, Fab{\'o},
  S{\'o}lyom, Er{\H o}ss, Telcs, and
  Somogyv{\'a}ri}]{benkoCompleteInferenceCausal2020}
Benk{\H o}, Z.; Zlatniczki, {\'A}.; Stippinger, M.; Fab{\'o}, D.; S{\'o}lyom,
  A.; Er{\H o}ss, L.; Telcs, A.; and Somogyv{\'a}ri, Z. 2020.
\newblock Complete {{Inference}} of {{Causal Relations}} between {{Dynamical
  Systems}}.
\newblock arxiv:1808.10806.

\bibitem[{Brouwer et~al.(2021)Brouwer, Arany, Simm, and
  Moreau}]{brouwerLatentConvergentCross2021}
Brouwer, E.~D.; Arany, A.; Simm, J.; and Moreau, Y. 2021.
\newblock Latent {{Convergent Cross Mapping}}.
\newblock In \emph{International {{Conference}} on {{Learning
  Representations}}}.

\bibitem[{Cheng et~al.(2023)Cheng, Yang, Xiao, Li, Suo, He, and
  Dai}]{chengCUTSNeuralCausal2023}
Cheng, Y.; Yang, R.; Xiao, T.; Li, Z.; Suo, J.; He, K.; and Dai, Q. 2023.
\newblock {{CUTS}}: {{Neural Causal Discovery}} from {{Irregular Time-Series
  Data}}.
\newblock In \emph{The {{Eleventh International Conference}} on {{Learning
  Representations}}}.

\bibitem[{Cho et~al.(2014)Cho, {van Merrienboer}, Gulcehre, Bahdanau, Bougares,
  Schwenk, and Bengio}]{choLearningPhraseRepresentations2014}
Cho, K.; {van Merrienboer}, B.; Gulcehre, C.; Bahdanau, D.; Bougares, F.;
  Schwenk, H.; and Bengio, Y. 2014.
\newblock Learning {{Phrase Representations}} Using {{RNN Encoder-Decoder}} for
  {{Statistical Machine Translation}}.
\newblock arxiv:1406.1078.

\bibitem[{Cini, Marisca, and Alippi(2022)}]{ciniFillingApMultivariate2022}
Cini, A.; Marisca, I.; and Alippi, C. 2022.
\newblock Filling the {{G}}\_ap\_s: Multivariate Time Series Imputation by
  Graph Neural Networks.
\newblock In \emph{International {{Conference}} on {{Learning
  Representations}}}.

\bibitem[{Cundy, Grover, and Ermon(2021)}]{cundyBCDNetsScalable2021}
Cundy, C.; Grover, A.; and Ermon, S. 2021.
\newblock {{BCD Nets}}: {{Scalable Variational Approaches}} for {{Bayesian
  Causal Discovery}}.
\newblock In \emph{Advances in {{Neural Information Processing Systems}}},
  volume~34, 7095--7110. {Curran Associates, Inc.}

\bibitem[{Fleuret and Geman(2001)}]{fleuretCoarsetoFineFaceDetection2001}
Fleuret, F.; and Geman, D. 2001.
\newblock Coarse-to-{{Fine Face Detection}}.
\newblock \emph{International Journal of Computer Vision}, 41(1): 85--107.

\bibitem[{Geffner et~al.(2022)Geffner, Antoran, Foster, Gong, Ma, Kiciman,
  Sharma, Lamb, Kukla, Pawlowski, Allamanis, and
  Zhang}]{geffnerDeepEndtoendCausal2022}
Geffner, T.; Antoran, J.; Foster, A.; Gong, W.; Ma, C.; Kiciman, E.; Sharma,
  A.; Lamb, A.; Kukla, M.; Pawlowski, N.; Allamanis, M.; and Zhang, C. 2022.
\newblock Deep {{End-to-end Causal Inference}}.
\newblock arxiv:2202.02195.

\bibitem[{Gerhardus and Runge(2020)}]{gerhardusHighrecallCausalDiscovery2020}
Gerhardus, A.; and Runge, J. 2020.
\newblock High-Recall Causal Discovery for Autocorrelated Time Series with
  Latent Confounders.
\newblock In \emph{Advances in {{Neural Information Processing Systems}}},
  volume~33, 12615--12625. {Curran Associates, Inc.}

\bibitem[{Gilmer et~al.(2017)Gilmer, Schoenholz, Riley, Vinyals, and
  Dahl}]{gilmerNeuralMessagePassing2017}
Gilmer, J.; Schoenholz, S.~S.; Riley, P.~F.; Vinyals, O.; and Dahl, G.~E. 2017.
\newblock Neural {{Message Passing}} for {{Quantum Chemistry}}.
\newblock In \emph{Proceedings of the 34th {{International Conference}} on
  {{Machine Learning}}}, 1263--1272. {PMLR}.

\bibitem[{Granger(1969)}]{grangerInvestigatingCausalRelations1969}
Granger, C. W.~J. 1969.
\newblock Investigating Causal Relations by Econometric Models and
  Cross-Spectral Methods.
\newblock \emph{Econometrica}, 37(3): 424--438.

\bibitem[{Hong, Liu, and Mai(2017)}]{hongEfficientAlgorithmLargescale2017}
Hong, Y.; Liu, Z.; and Mai, G. 2017.
\newblock An Efficient Algorithm for Large-Scale Causal Discovery.
\newblock \emph{Soft Computing}, 21(24): 7381--7391.

\bibitem[{Hoyer et~al.(2008)Hoyer, Janzing, Mooij, Peters, and
  Sch{\"o}lkopf}]{hoyerNonlinearCausalDiscovery2008}
Hoyer, P.; Janzing, D.; Mooij, J.~M.; Peters, J.; and Sch{\"o}lkopf, B. 2008.
\newblock Nonlinear Causal Discovery with Additive Noise Models.
\newblock In \emph{Advances in {{Neural Information Processing Systems}}},
  volume~21. {Curran Associates, Inc.}

\bibitem[{Jang, Gu, and
  Poole(2016)}]{jangCategoricalReparameterizationGumbelsoftmax2016}
Jang, E.; Gu, S.; and Poole, B. 2016.
\newblock Categorical Reparameterization with Gumbel-Softmax.

\bibitem[{Karimi and Paul(2010)}]{karimiExtensiveChaosLorenz962010}
Karimi, A.; and Paul, M.~R. 2010.
\newblock Extensive Chaos in the Lorenz-96 Model.
\newblock \emph{Chaos: An Interdisciplinary Journal of Nonlinear Science},
  20(4): 043105.

\bibitem[{Khanna and Tan(2020)}]{khannaEconomyStatisticalRecurrent2020}
Khanna, S.; and Tan, V. Y.~F. 2020.
\newblock Economy Statistical Recurrent Units for Inferring Nonlinear Granger
  Causality.
\newblock In \emph{International {{Conference}} on {{Learning
  Representations}}}.

\bibitem[{Lopez et~al.(2022)Lopez, Huetter, Pritchard, and
  Regev}]{lopezLargeScaleDifferentiableCausal2022a}
Lopez, R.; Huetter, J.-C.; Pritchard, J.; and Regev, A. 2022.
\newblock Large-{{Scale Differentiable Causal Discovery}} of {{Factor Graphs}}.
\newblock \emph{Advances in Neural Information Processing Systems}, 35:
  19290--19303.

\bibitem[{L{\"o}we et~al.(2022)L{\"o}we, Madras, Zemel, and
  Welling}]{loweAmortizedCausalDiscovery2022}
L{\"o}we, S.; Madras, D.; Zemel, R.; and Welling, M. 2022.
\newblock Amortized Causal Discovery: Learning to Infer Causal Graphs from
  Time-Series Data.
\newblock In \emph{Proceedings of the {{First Conference}} on {{Causal
  Learning}} and {{Reasoning}}}, 509--525. {PMLR}.

\bibitem[{{Morales-Alvarez} et~al.(2022){Morales-Alvarez}, Gong, Lamb,
  Woodhead, Peyton~Jones, Pawlowski, Allamanis, and
  Zhang}]{morales-alvarezSimultaneousMissingValue2022}
{Morales-Alvarez}, P.; Gong, W.; Lamb, A.; Woodhead, S.; Peyton~Jones, S.;
  Pawlowski, N.; Allamanis, M.; and Zhang, C. 2022.
\newblock Simultaneous {{Missing Value Imputation}} and {{Structure Learning}}
  with {{Groups}}.
\newblock \emph{Advances in Neural Information Processing Systems}, 35:
  20011--20024.

\bibitem[{Pamfil et~al.(2020)Pamfil, Sriwattanaworachai, Desai, Pilgerstorfer,
  Georgatzis, Beaumont, and Aragam}]{pamfilDYNOTEARSStructureLearning2020}
Pamfil, R.; Sriwattanaworachai, N.; Desai, S.; Pilgerstorfer, P.; Georgatzis,
  K.; Beaumont, P.; and Aragam, B. 2020.
\newblock {{DYNOTEARS}}: Structure Learning from Time-Series Data.
\newblock In \emph{Proceedings of the {{Twenty Third International Conference}}
  on {{Artificial Intelligence}} and {{Statistics}}}, 1595--1605. {PMLR}.

\bibitem[{Peters, Janzing, and
  Sch{\"o}lkopf(2017)}]{petersElementsCausalInference2017}
Peters, J.; Janzing, D.; and Sch{\"o}lkopf, B. 2017.
\newblock \emph{Elements of Causal Inference: Foundations and Learning
  Algorithms}.
\newblock {The MIT Press}.
\newblock ISBN 978-0-262-03731-0 978-0-262-34429-6.

\bibitem[{Prill et~al.(2010)Prill, Marbach, {Saez-Rodriguez}, Sorger,
  Alexopoulos, Xue, Clarke, {Altan-Bonnet}, and
  Stolovitzky}]{prillRigorousAssessmentSystems2010}
Prill, R.~J.; Marbach, D.; {Saez-Rodriguez}, J.; Sorger, P.~K.; Alexopoulos,
  L.~G.; Xue, X.; Clarke, N.~D.; {Altan-Bonnet}, G.; and Stolovitzky, G. 2010.
\newblock Towards a {{Rigorous Assessment}} of {{Systems Biology Models}}:
  {{The DREAM3 Challenges}}.
\newblock \emph{PLOS ONE}, 5(2): e9202.

\bibitem[{Runge(2020)}]{rungeDiscoveringContemporaneousLagged2020}
Runge, J. 2020.
\newblock Discovering Contemporaneous and Lagged Causal Relations in
  Autocorrelated Nonlinear Time Series Datasets.
\newblock In \emph{Proceedings of the 36th {{Conference}} on {{Uncertainty}} in
  {{Artificial Intelligence}} ({{UAI}})}, 1388--1397. {PMLR}.

\bibitem[{Runge et~al.(2019)Runge, Nowack, Kretschmer, Flaxman, and
  Sejdinovic}]{rungeDetectingQuantifyingCausal2019}
Runge, J.; Nowack, P.; Kretschmer, M.; Flaxman, S.; and Sejdinovic, D. 2019.
\newblock Detecting and Quantifying Causal Associations in Large Nonlinear Time
  Series Datasets.
\newblock \emph{Science Advances}, 5(11): eaau4996.

\bibitem[{Sarlin et~al.(2019)Sarlin, Cadena, Siegwart, and
  Dymczyk}]{sarlinCoarseFineRobust2019}
Sarlin, P.-E.; Cadena, C.; Siegwart, R.; and Dymczyk, M. 2019.
\newblock From {{Coarse}} to {{Fine}}: {{Robust Hierarchical Localization}} at
  {{Large Scale}}.
\newblock In \emph{Proceedings of the {{IEEE}}/{{CVF Conference}} on {{Computer
  Vision}} and {{Pattern Recognition}}}, 12716--12725.

\bibitem[{Shimizu et~al.(2006)Shimizu, Hoyer, Hyv\&\#228, {rinen}, and
  Kerminen}]{shimizuLinearNonGaussianAcyclic2006}
Shimizu, S.; Hoyer, P.~O.; Hyv\&\#228, A.; {rinen}; and Kerminen, A. 2006.
\newblock A {{Linear Non-Gaussian Acyclic Model}} for {{Causal Discovery}}.
\newblock \emph{Journal of Machine Learning Research}, 7(72): 2003--2030.

\bibitem[{Spirtes and Glymour(1991)}]{spirtesAlgorithmFastRecovery1991}
Spirtes, P.; and Glymour, C. 1991.
\newblock An Algorithm for Fast Recovery of Sparse Causal Graphs.
\newblock \emph{Social science computer review}, 9(1): 62--72.

\bibitem[{Spirtes et~al.(2000)Spirtes, Glymour, Scheines, and
  Heckerman}]{spirtesCausationPredictionSearch2000}
Spirtes, P.; Glymour, C.~N.; Scheines, R.; and Heckerman, D. 2000.
\newblock \emph{Causation, Prediction, and Search}.
\newblock {MIT press}.

\bibitem[{Sugihara et~al.(2012)Sugihara, May, Ye, Hsieh, Deyle, Fogarty, and
  Munch}]{sugiharaDetectingCausalityComplex2012}
Sugihara, G.; May, R.; Ye, H.; Hsieh, C.-h.; Deyle, E.; Fogarty, M.; and Munch,
  S. 2012.
\newblock Detecting {{Causality}} in {{Complex Ecosystems}}.
\newblock \emph{Science}, 338(6106): 496--500.

\bibitem[{Tank et~al.(2022)Tank, Covert, Foti, Shojaie, and
  Fox}]{tankNeuralGrangerCausality2022}
Tank, A.; Covert, I.; Foti, N.; Shojaie, A.; and Fox, E.~B. 2022.
\newblock Neural Granger Causality.
\newblock \emph{IEEE Transactions on Pattern Analysis and Machine
  Intelligence}, 44(8): 4267--4279.

\bibitem[{Vowels, Camgoz, and Bowden(2021)}]{vowelsYaDAGsSurvey2021}
Vowels, M.~J.; Camgoz, N.~C.; and Bowden, R. 2021.
\newblock D'ya like Dags? {{A}} Survey on Structure Learning and Causal
  Discovery.
\newblock arxiv:2103.02582.

\bibitem[{Wu, Singh, and Berger(2022)}]{wuGrangerCausalInference2022}
Wu, A.~P.; Singh, R.; and Berger, B. 2022.
\newblock Granger Causal Inference on Dags Identifies Genomic Loci Regulating
  Transcription.
\newblock In \emph{International {{Conference}} on {{Learning
  Representations}}}.

\bibitem[{Wu et~al.(2023)Wu, Hu, Liu, Zhou, Wang, and
  Long}]{wuTimesNetTemporal2DVariation2023}
Wu, H.; Hu, T.; Liu, Y.; Zhou, H.; Wang, J.; and Long, M. 2023.
\newblock {{TimesNet}}: {{Temporal 2D-Variation Modeling}} for {{General Time
  Series Analysis}}.
\newblock In \emph{The {{Eleventh International Conference}} on {{Learning
  Representations}}}.

\bibitem[{Xu, Huang, and Yoo(2019)}]{xuScalableCausalGraph2019}
Xu, C.; Huang, H.; and Yoo, S. 2019.
\newblock Scalable {{Causal Graph Learning}} through a {{Deep Neural Network}}.
\newblock In \emph{Proceedings of the 28th {{ACM International Conference}} on
  {{Information}} and {{Knowledge Management}}}, {{CIKM}} '19, 1853--1862. {New
  York, NY, USA}: {Association for Computing Machinery}.
\newblock ISBN 978-1-4503-6976-3.

\bibitem[{Ye et~al.(2015)Ye, Deyle, Gilarranz, and
  Sugihara}]{yeDistinguishingTimedelayedCausal2015}
Ye, H.; Deyle, E.~R.; Gilarranz, L.~J.; and Sugihara, G. 2015.
\newblock Distinguishing Time-Delayed Causal Interactions Using Convergent
  Cross Mapping.
\newblock \emph{Scientific Reports}, 5(1): 14750.

\bibitem[{Yi et~al.(2016)Yi, Zheng, Zhang, and Li}]{yiSTMVLFillingMissing2016}
Yi, X.; Zheng, Y.; Zhang, J.; and Li, T. 2016.
\newblock {{ST-MVL}}: Filling Missing Values in Geo-Sensory Time Series Data.
\newblock In \emph{Proceedings of the 25th {{International Joint Conference}}
  on {{Artificial Intelligence}}}.

\end{thebibliography}

\end{document}